\documentclass[journal]{IEEEtran}
\usepackage{graphicx}
\usepackage{amsmath}
\usepackage{times}
\usepackage{epsfig}
\usepackage{amssymb}
\usepackage{multirow}
\usepackage{color}
\usepackage{eso-pic}
\usepackage{url}
\usepackage{xspace}
\usepackage[breaklinks=true,bookmarks=false]{hyperref}
\usepackage{float}
\usepackage{bbding}
\usepackage{pifont}
\usepackage{wasysym}

\makeatletter
\DeclareRobustCommand\onedot{\futurelet\@let@token\@onedot}
\def\@onedot{\ifx\@let@token.\else.\null\fi\xspace}

\def\eg{\emph{e.g}\onedot} 
\def\ie{\emph{i.e}\onedot} 
 
\def\etc{\emph{etc}\onedot} 
 
\def\etal{\emph{et al}\onedot}
\makeatother

\ifCLASSINFOpdf
\else
\fi

\begin{document}
%
\title{A Robust {\color{black}Attentional} Framework for License Plate Recognition in the Wild}
%
%
%

\author{ Linjiang Zhang$^{\star}$, Peng Wang$^{\star}$, Hui Li$^{\dagger}$, Zhen Li, Chunhua Shen, Yanning Zhang
    \thanks{L. Zhang, P. Wang and Y. Zhang are with the School of Computer Science, Northwestern Polytechnical University, Xi'an, China, and National Engineering Laboratory for Integrated Aero-Space-Ground-Ocean Big Data Application Technology, China. H. Li and C. Shen are with the School of Computer Science, The University of Adelaide, South Australia, 5005, Australia. Z. Li is with the big data and AI technology department at Minsheng Fintech Corp. LTD.
    $^{\dagger}$H. Li is the corresponding author (E-mail: huili03855@gmail.com).
    $^{\star}$L. Zhang and P. Wang contribute equally to this work.
    This work is supported in part by the National Natural Science Foundation of China (No. 61876152) and the seed Foundation of Innovation and Creation for Graduate Students in Northwestern Polytechnical University (No. CX2020184).}
}

\markboth{}%
{Wang \MakeLowercase{\textit{et al.}}:
	 A Simple and Robust Network for License Plate Recognition in the Wild}

\maketitle

\begin{abstract}
Recognizing car license plates in natural scene images is an important yet still challenging task in realistic applications. 
Many existing approaches perform well for license plates collected under constrained conditions, \eg, 
shooting in frontal and horizontal view-angles and under good lighting conditions.
However, their performance drops significantly in an unconstrained environment that features rotation, distortion, occlusion, blurring, shading or extreme dark or bright conditions. 
In this work, we propose a robust framework for license plate recognition in the wild. It is composed of a tailored CycleGAN model for license plate image generation and an elaborate designed image-to-sequence network for plate recognition. On one hand, the CycleGAN based plate generation engine alleviates the exhausting human annotation work. Massive amount of training data can be obtained with a more balanced character distribution and various shooting conditions, which helps to boost the recognition accuracy to a large extent. On the other hand, the 2D attentional based license plate recognizer with an Xception-based CNN encoder is capable of recognizing license plates with different patterns under various scenarios accurately and robustly. Without using any heuristics rule or post-processing, our method achieves the state-of-the-art performance on four public datasets, which demonstrates the generality and robustness of our framework. Moreover, we released a new license plate dataset, named ``CLPD'', with $1200$ images from all $31$ provinces in mainland China. The dataset can be available from: https://github.com/wangpengnorman/CLPD\_dataset.
\end{abstract}


%
\IEEEpeerreviewmaketitle

\section{Introduction}

\IEEEPARstart{L}{icense} Plate (LP) recognition in the wild is a fundamental problem in intelligent transportation systems. It can be used in a variety of applications including self-driving vehicles, traffic control and surveillance. 

\begin{figure}[!t]
\centering
\includegraphics[scale=0.32]{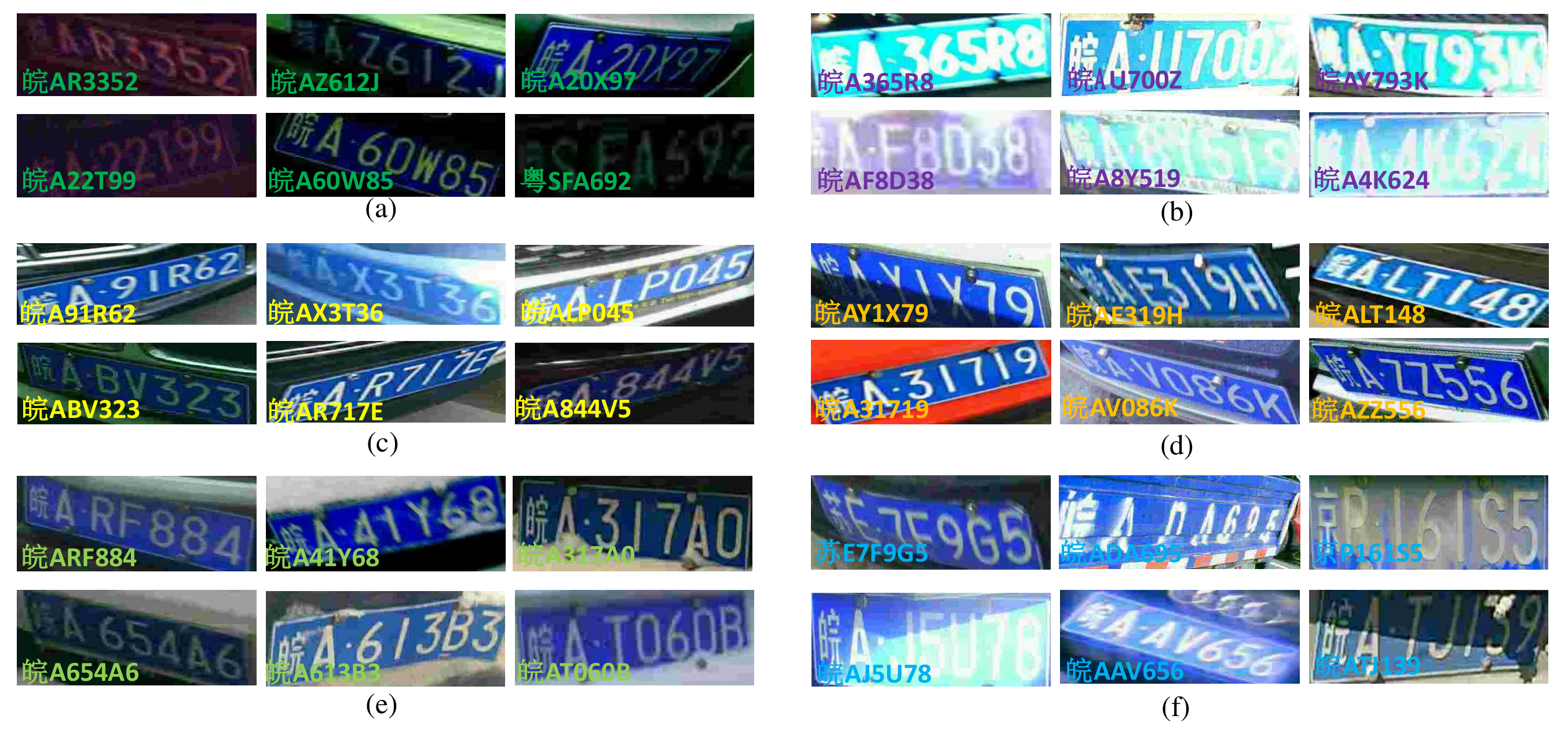}
\vspace{-8mm}
\caption{
Examples of license plates successfully recognized by our proposed algorithm.
(a) Dark Illumination; (b) Extremely bright or uneven; (c) Large horizontal tilt degree; (d) Large vertical tilt degree; (e) Images taken on a snowy or rainy day; (f) Mixture of bad conditions.} 
\label{CCPD}
\end{figure}

The LP numbers enable the link to a large body of information, including ownership, vehicle condition and driving record.
Therefore, the technique of LP recognition in the wild can play a key role in road safety, traffic control and law enforcement.
Although the recognition accuracy is acceptable for images shot under constrained conditions, recognizing license plates in complex environment is still far from satisfactory, especially for images photographed in dark, glare, occluded, rainy, snowy, tilted or blurred scenarios as shown in Figure~\ref{CCPD}.

With the advantage of deep neural networks, numerous work is proposed in recent years for license plate recognition, with Convolutional Neural Networks (CNNs) used for feature extraction, and Connectionist Temporal Classification (CTC)~\cite{LiReading2016}, number classifiers~\cite{Xu2018Towards}, \etc followed for character reading. These methods perform well for regular license plates (\eg, nearly horizontal). When the license plate images are tilted or bent, an extra rectification step is required before recognition~\cite{Silva2018License}.

This paper tackles the task of license plate recognition in unconstrained scenarios. A robust framework is proposed to handle license plate recognition in both regular and challenging cases effectively. Our proposed license plate recognizer is composed of a 30-layer lightweight Xception for feature extraction and a 2D-attention based decoding module for character sequence recognition. Without extra processings like image rectification or character segmentation, the proposed model is capable of recognizing license plates in both regular and irregular patterns under various practical scenarios.
Different from current methods of treating a license plate as a one-dimensional sequence, our method uses 2D-attention that considers license plate image as a 2-dimensional signal. Trained in a weakly supervised manner, the proposed model is able to approximately localize the corresponding characters on license plates in decoding process, regardless of the appearance of license plate patterns. 

Many license plate datasets are collected from one region, which causes bias in the datasets. For example, Xu \textsl{et al.} \cite{Xu2018Towards} introduce a license plate dataset \textbf{CCPD} which contains about 290K real world license plate images in various complex situations, as shown in Figure \ref{CCPD}.  However, since more than $95\%$ of the images are photographed in one city, the first two characters in license plates are mostly the same, which may lead to bias for the trained model. In order to obtain a robust model which can be generally used for recognizing license plates from different regions, a CycleGAN model is tailored here which can mimic real scenarios and generate different kinds of license plate images, such as in dark or strong lighting conditions, containing shadows, \etc. Moreover, license plates with various province characters can be synthesized, which alleviates the exhausting human annotation work to a large extent and enables a more general license plate recognition model. Our framework is evaluated on four public datasets. The competitive performance demonstrates the robustness of our framework. Moreover, we also collect a new license plate dataset with images from all $31$ provinces in China, named ``\textbf{CLPD}''. It enables a more comprehensive evaluation of current plate recognition methods, and promotes the research of a more practical model.

It should be noted that the focus of this work is license plate recognition. So we simply train the off-the-shelf YOLOv2 detector~\cite{DBLP:journals/corr/RedmonF16} here to obtain bounding boxes of license plates. 

The main contributions of this paper can be summarized as follows:

1. We design a robust method for license plate recognition in natural scene images. It is made up of a tailored Xception module and an encoder-decoder module. We optimized the recognition framework by using a 2D attention mechanism.  It is able to extract local features for individual characters in a weakly supervised manner, without character level annotations needed. Compared to existing license plate recognition approaches, our method does not need an extra module to handle the irregularity of license plates or segment each character for recognition.


2. A tailored CycleGAN is proposed to synthesize license plates under various scenarios, including adding shadows, glare or darkness, perspective transformation, \etc.. With this engine we can generate license plate images with less data bias, and so get models with better generalization abilities.

3. We build a new dataset, named CLPD.
It covers a large variety of photographing conditions, vehicle types and region codes, which provides a more comprehensive evaluation benchmark for plate recognition algorithms and promotes a more practical model design.

\begin{figure*}[t!]
\includegraphics[width=0.9\textwidth]{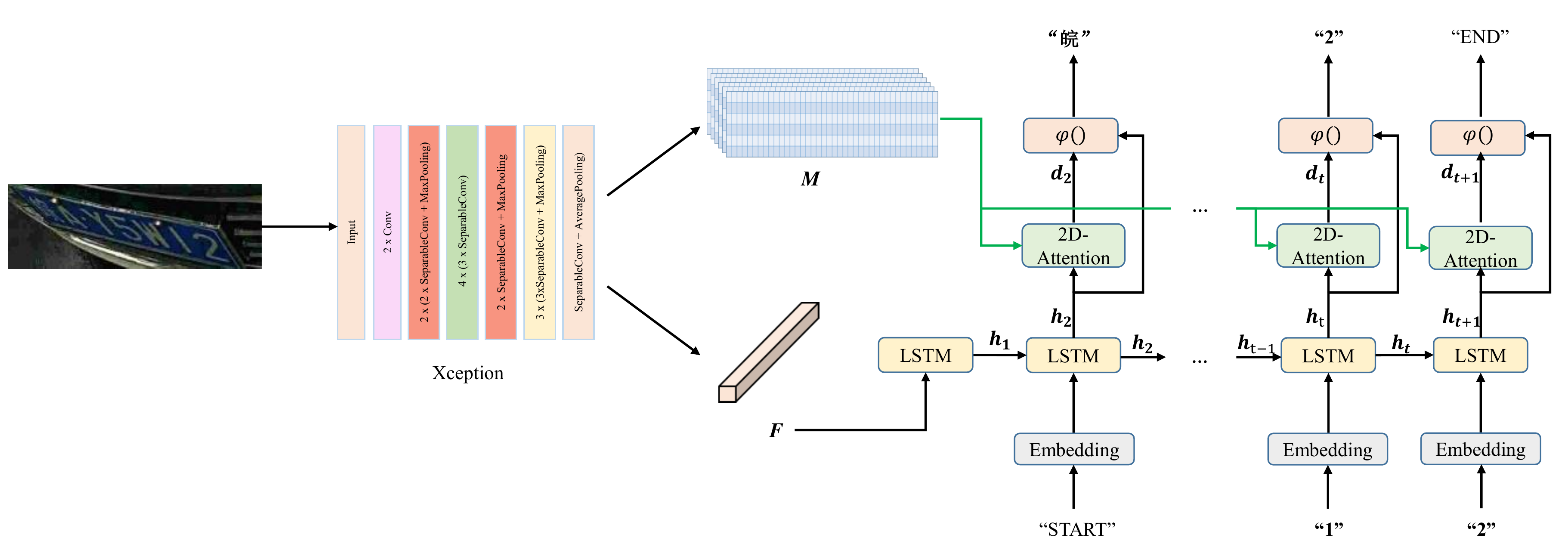}
\caption{Overview of the proposed architecture for LP recognition in complex scenarios. We extract license plates via a well-trained YOLOv2. The detected bounding box is fed into a 30-layer Xception network and get a global feature vector (denoted as $F$). An LSTM model is adopted to decode the obtained image feature into license plate numbers. 
We also extract an intermediate feature map (denoted as $M$) from the $12$th layer of Xception, which provides local features during character decoding process. }
\label{model}
\end{figure*}

\section{Related Work}
In this section, we present a concise introduction to related works on license plate recognition, light-weight convolutional neural networks, generative adversarial networks and datasets of license plate.

\subsection{License Plate Recognition}
Existing methods for license plate recognition can be divided into two categories: Segmentation based~\cite{Silva2018License,Gou2016Vehicle,Guo2008License,Gonalves2016,Li2018Rotation} and Non-segmentation based methods~\cite{LiReading2016,Xu2018Towards,Hui2017Toward}. The segmentation based methods generally segment the license plate into characters and then recognize individual characters by OCR models~\cite{Silva2018License,gou2015vehicle,bulan2017segmentation}. 
Bulan~\etal~\cite{bulan2017segmentation} perform segmentation and OCR jointly by using a hidden Markov models (HMMs) based probabilistic inference method, where the most likely character sequence is determined by Viterbi algorithm. Segmentation based methods rely heavily on the segmentation performance, which is very susceptible to the environment, including strong or weak lighting, bad weather, blurring, \etc, and will result in a low recognition accuracy even with a strong recognizer. 

Recent methods are mostly segmentation free. For example, Li~\etal~\cite{LiReading2016} propose to treat license plate as a character sequence. Sequential features are encoded by CNNs and Bidirectional RNNs (BRNNs), and decoded by CTC without character separation. The CNN features are extracted from a well-trained CNN classifier, and the model cannot be trained end-to-end. RPnet proposed by Xu~\etal~\cite{Xu2018Towards} extracts ROI features from several different convolutional layers, and feeds the combined feathers to a series of classifiers for recognition. The number of classifiers is determined by the number of characters in the license plate, which limits its generalization ability in different regions. Li~\etal~\cite{Hui2017Toward} later propose a unified network which is able to localize license plates and recognize the letters at the same time in a single forward process. Similarly, the region features are encoded by BRNNs and decoded by CTC, which restricts its application to oriented LPs. Compared to the previous work, our method uses a 2D attention based encoder-decoder framework, where characters can be approximately localized by 2D attention regardless of LP image appearance, which enables its application to arbitrarily-oriented LPs.


\subsection{Scene Text Recognition}

License plate recognition can be regarded as a special case of general scene text recognition tasks, which have different characteristics. Characters in license plate usually use the same font in one region. There is no language model hidden in license plate, and no strong relationship with the context semantic information. In contrast, general scene text has a great variability on fonts. A language lexicon is existed and the text content is often highly relevant to the objects or scenes of the image.
Xie~\etal~\cite{xie2019aggregation} propose a novel method where aggregation cross-entropy (ACE) is used for sequence recognition, replacing the generally used CTC loss owing to its inconvenience in processing 2D problems.  A multi-object rectified attention network (MORAN) for scene text recognition is proposed by Luo~\etal~\cite{luo2019moran}, which contains a multi-object rectification network (MORN) and an attention-based sequence recognition network (ASRN). The image is rectified by MORN and then input to ASRN for recognition. Shi~\etal~\cite{shi2018aster} put forward a system that a flexible Thin-Plate Spline transformation is used to adaptively rectify a text image. A recognition model predicts a character sequence directly from the rectified image. Li~\etal~\cite{li2019show} use a 2D attention based encoder-decoder framework for irregular text recognition, which is very similar to our work. However, in our framework, a tailored CycleGAN is added for synthetic license plate generation, which can reduce data bias and improve model generalization ability.


\subsection{Generative Adversarial Networks}
With the invention of  Generative Adversarial Networks (GANs)~\cite{Goodfellow2014Generative}, many improved models have emerged, such as Deep Convolutional GANs (DCGANs) \cite{Radford2015Unsupervised}, Conditional GAN \cite{Mirza2014Conditional}, Cycle-Consistent Adversarial Networks (CycleGAN)~\cite{Zhu2017Unpaired}, Wasserstein GANs (WGAN) \cite{Arjovsky2017Wasserstein} \etc.. Zhu~\etal~\cite{Zhu2017Unpaired} propose the CycleGAN, which learns the mapping between an input image and an output image using a training set of unaligned image pairs. In order to migrate the style of one image set to another one, cycle consistency loss is introduced. Based on this model, we propose an improved algorithm to generate synthetic license plate images in more complex environments, which improves the accuracy of license plate recognition furthermore. Wang~\etal~\cite{Wang2017Adversarial} adopt CycleWGAN to generate license plate images for improving recognition performance. Images simulating different shotting conditions are generated simultaneously. BRNN+CTC is used for plate recognition, which does not take oriented license plates into consideration as well. Nevertheless, we use a tailored CycleGAN to generate license plates under different conditions separately, which can lead to a better recognition performance.

\subsection{Datasets of License Plate}
Most datasets about license plates detection and recognition are collected from one area, and the type of license plate is monotonous (\eg, only containing civic cars, no buses or trucks). Images are taken under similar conditions, such as highway toll stations and parking lots. Hence those datasets could not verify the robustness of a model.

Silva~\etal~\cite{Silva2018License} collect a dataset named CD-HARD with $102$ images, which covers some difficult situations, including tilting. However, because of the small number of images, the test result is susceptible to tricks. PKUData~\cite{PKUData} captures images through a road surveillance camera, which includes a variety of license plate types and different lighting conditions. Unfortunately, all license plates are horizontal and taken from one province which has the same province code. Models trained on PKUData cannot be used to recognize license plates from other regions.  AOLP~\cite{AOLP} database consists of $2049$ images with Taiwan license plate. This dataset is categorized into three subsets according to different levels of difficulty and photographing conditions. 
CCPD is currently the largest license plate dataset with $290$k images, and is divided into multiple subsets such as tilt, difficulty, glare, and distance according to license plate conditions, which contributes greatly to the community. Nevertheless, more than $90\%$ of the images are from one city too, which limits the trained model to recognize license plates from other areas. In this work, we propose to synthesize license plates by CycleGAN so as to make up for the deficiency. A new dataset names CLPD is introduced, which includes license plates from different provinces, to evaluate recognition models comprehensively.

\begin{figure*}[ht!]
\centering
\includegraphics[width=0.8\textwidth]{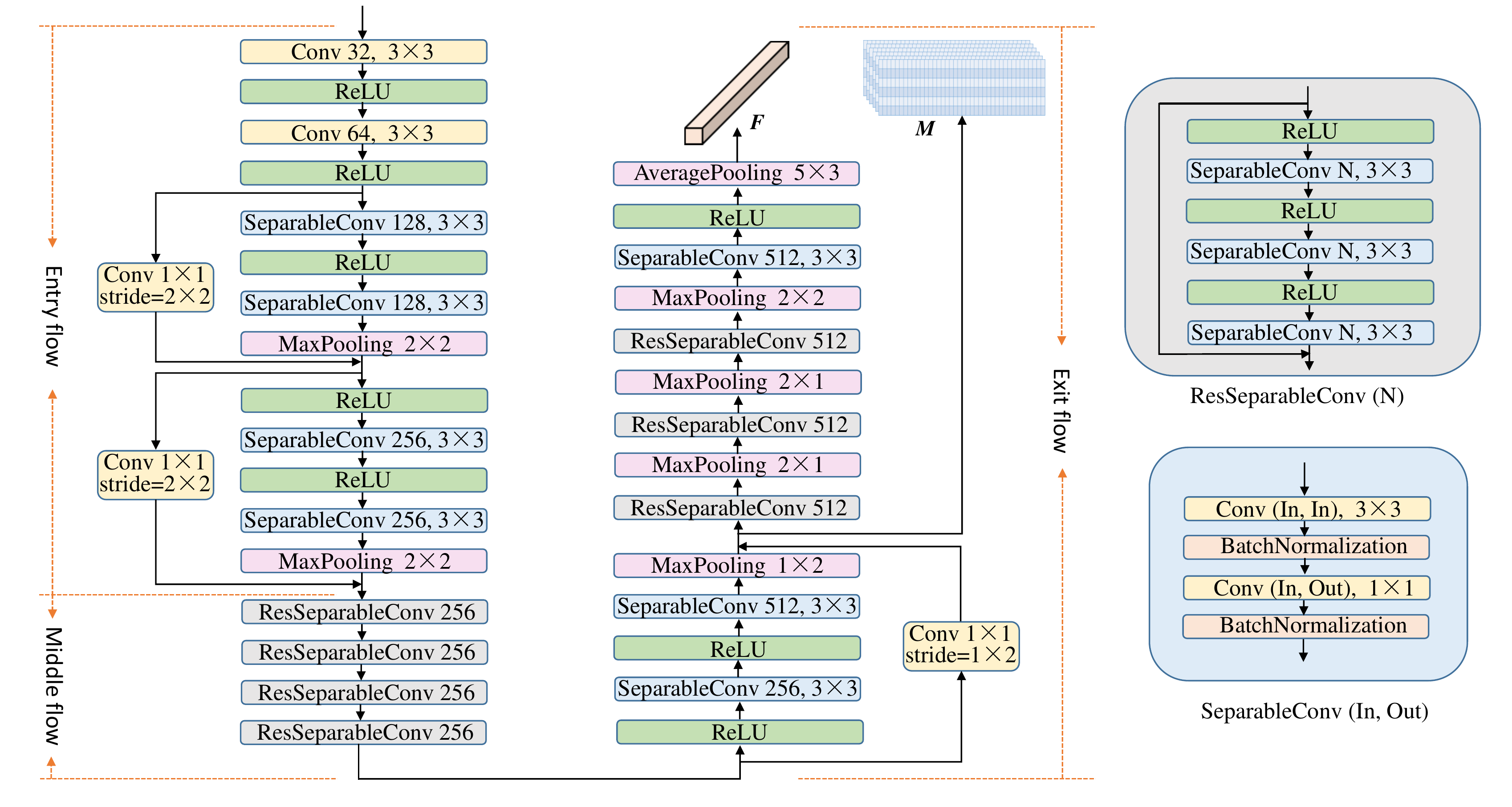}
\caption{The tailored Xception architecture. ``Conv'' stands for Convolutional layers, with output channels and kernel sizes presented. The stride and padding for convolutional layers are all set to $1$, and no padding for Max-pooling layers.} 
\label{xception}
\end{figure*}

\section{Model}
We introduce our proposed model in this section. As presented in Figure \ref{model}, the whole LP recognition model consists of two main parts: a tailored Xception network for feature extraction and a 2D-attention based RNN model for character decoding.

\subsection{The Convolutional Image Encoder}
A 30-layer Xception encoder is tailored from the original Xception~\cite{Chollet2017Xception} framework to fit our application, whose details are presented in Figure~\ref{xception}. The convolutional parts of our model are based entirely on depthwise separable convolution layers~\cite{sifre2014rigid}. The $30$ convolutional layers are structured into $9$ modules, where all of them have linear residual connections except for the first and the last one. The term ``ResSeparableConv'' stands for a stack of three separable convolution layers with an identity residual connection.

The entry flow downsamples the spatial size from $160\times48$ to $40\times6$ and increases the feature channel from $3$ to $256$ using interleaved separable convolutions and max-poolings.
In the middle flow, we adopt repeated ResSeparableConv blocks to extract deep features 
that contain higher level representations, while the spatial size and channel number are fixed.
In the exit flow, we extract a middle-level feature map $M$ of size $40$$\times$$6$$\times$$512$ as context for attention network and a final feature vector $F$ of $512$ dimensions.


\subsection{The Recurrent Sequence Decoder}
RNN is widely used in translation, image caption, scene text recognition tasks. Here we extend it to license plate recognition. With a two-dimensional attention mechanism integrated, there is no need to make corrections for irregular license plate images or segment out each character for recognition. The proposed model can handle LPs in arbitrary shapes.

\begin{figure}[t]
\centering
\includegraphics[scale=0.58]{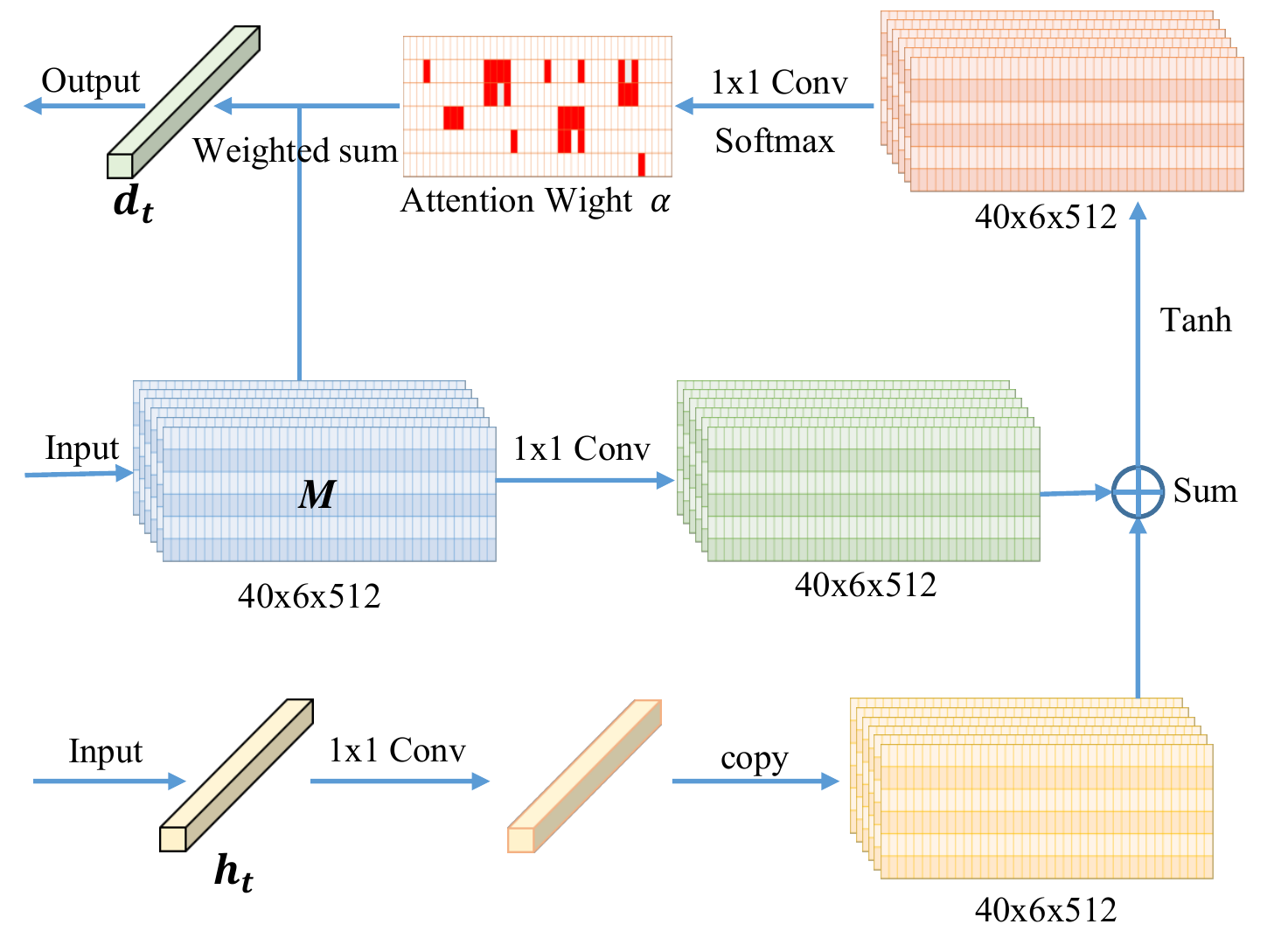}
\caption{The schematic of the 2D attention mechanism. $M$ is the feature map of the image obtained by Xception (as shown in Figure~\ref{xception}), and $h_{t}$ is the hidden state of each time step in decoding.} 
\label{2Dattention}
\end{figure}

$2$-layer LSTMs with $512$ hidden states each are adopted here in the sequence decoder. As shown in Figure~\ref{model}, the holistic feature vector $F$ is fed into LSTMs at time step $0$, which aims to provide an overall information about the input image. Then a ``START'' token is input into the model at time step $1$. From time step $2$, the output of previous time step is fed into LSTMs until the ``END'' token received. The inputs of LSTMs are embedded by one-hot vectors followed by a linear transformation. 
The calculation of a single LSTM cell in training can be expressed as:
\begin{equation}
h_{t+1} = \mathrm{f}(h_{t}, \psi(x_{t})), \qquad t = 1,...,8.
\end{equation}
where $h_{t}$ is the current hidden state, $\mathrm{f}()$ represents the LSTM operation at each time step and $\psi(\cdot)$ is the embedding operation. In inferring process, $x_{t} = \text{softmax} (\varphi(h_{t}, d_{t}) )$ which is the current output, while in training stage, the groundtruth character is adopted directly as $x_t$. $\varphi(\cdot)$ is a linear transformation, and $d_{t}$ is the output of the 2D-attention module, which is calculated as follows: 

\begin{equation}
\begin{cases}
g_{ij} = \mathrm{tanh}(W_{m}M_{ij} + W_{h}h_{t}), \,\, \\ 
\alpha_{ij} = \mathrm{softmax}(W_{g} \cdot g_{ij}), \\
d_{t} =  \sum_{i=1}^{H} \sum_{j=1}^{W} \alpha_{ij}M_{ij}
\end{cases}
\label{eq:atten}
\end{equation}
where $M_{ij}$ is the feature vector at position ${i, j}$ in $M$ and $h_{t}$ is the hidden state at time step $t$. $W_{m}, W_{h}, W_{g}$ are linear transformation matrices to be learned; $\alpha_{ij}$ is the attention weight at location ${i, j}$; $d_{t}$ is the weighted sum of image features, i.e., the local feature of the characters to be decoded at current time step $t$. The schematic of the 2D attention mechanism is illustrated in Figure~\ref{2Dattention}.

\section{AsymCycleGAN for LP Image Generation}
As aforementioned, it is difficult to manually collect LP images from a variety of regions,
which makes most existing LP datasets heavily biased towards specific regional identifiers.
In this section, we introduce a method for generating high-quality synthetic LP images using OpenCV and a tailored CycleGAN model (termed as AsymCycleGAN).
With this approach, we are able to construct a balanced training data and reduce the reliance on manually collected data.

\subsection{The Architecture of AsymCycleGAN}
\begin{figure}[t]
\centering
\includegraphics[scale=0.22]{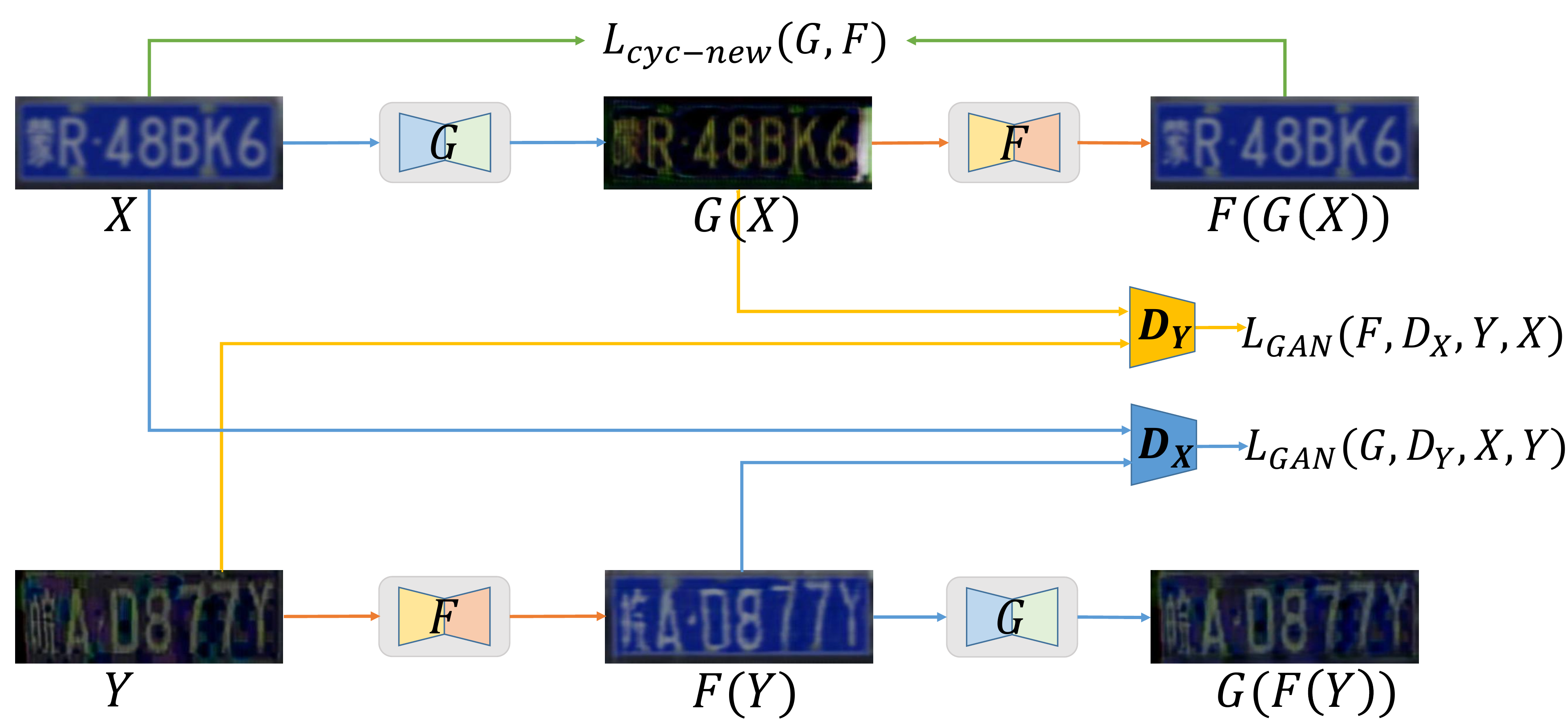}
\caption{The architecture of the proposed AsymCycleGAN model. $X$ are synthetic LP images generated by OpenCV,  $Y$ are real LP images.}
\label{CycleGAN}
\end{figure}

CycleGAN is an approach to translate an image from a source domain $X$ to a target domain $Y$ in the absence of paired training examples. 
In this work, the source domain $X$ is composed of fake LP images generated by OpenCV and 
the target domain $Y$ is made up of real LP images.
There are four learnable modules in CycleGAN, leading to two mapping functions $G: X \rightarrow Y, F: Y \rightarrow X$ and two discriminators
$D_{X}$ and $D_{Y}$.
The loss function of the standard CycleGAN can be expressed as follows:
\begin{equation}
\begin{aligned}
L(G,F,D_{X},D_{Y}) & =  L_{GAN}(G, D_{Y}, X, Y) \\
& +L_{GAN}(F,D_{X},Y, X) \\
& +\lambda L_{cyc}(G,F), 
\end{aligned}
\label{CycleGanLoss}
\end{equation}
where $L_{GAN}$ represents the adversarial loss and $L_{cyc}$ denotes the cycle-consistency loss:
\begin{equation}
\begin{aligned}
L_{cyc}(G,F) =  & E_{x \sim p_{data}(x)}[||F(G(x)) - x||_{1}] \\
 + & E_{y \sim p_{data}(y)}[||G(F(y)) - y||_{1}].
\label{eq:lcyc}
\end{aligned}
\end{equation}
In our case, what we need is the mapping function $G$ to generate real images from synthetic images. 
$F(G(x))$ can be roughly regarded as generating a noisy image from a clean one and then remove these noises, while $G(F(y))$ is the opposite process.
Note that in the process of $y \rightarrow F(y) \rightarrow G(F(y))$, 
the noise in $y$ removed by $F$ is in theory difficult to be exactly recovered by $G$, as one clean image can be associated to multiple real images with different noises. 
To this end, we replace the original cycle-consistency loss $L_{cyc}$~\eqref{eq:lcyc} with
\begin{equation}
\begin{aligned}
L_{cyc-new}(G,F) = E_{x \sim p_{data}(x)}[||F(G(x)) - x||_{1}],
\label{newCycle}
\end{aligned}
\end{equation}
where the term with respect to $G(F(y))$ is removed. 
We term the modified CycleGAN model is AsymCycleGAN, as its cycle-consistency loss is asymmetric. The architecture of the proposed AsymCycleGAN model is shown in Figure~\ref{CycleGAN}.

\subsection{AsymCycleGAN Generation Results}
\begin{figure}[!t]
\centering
\includegraphics[scale=0.54]{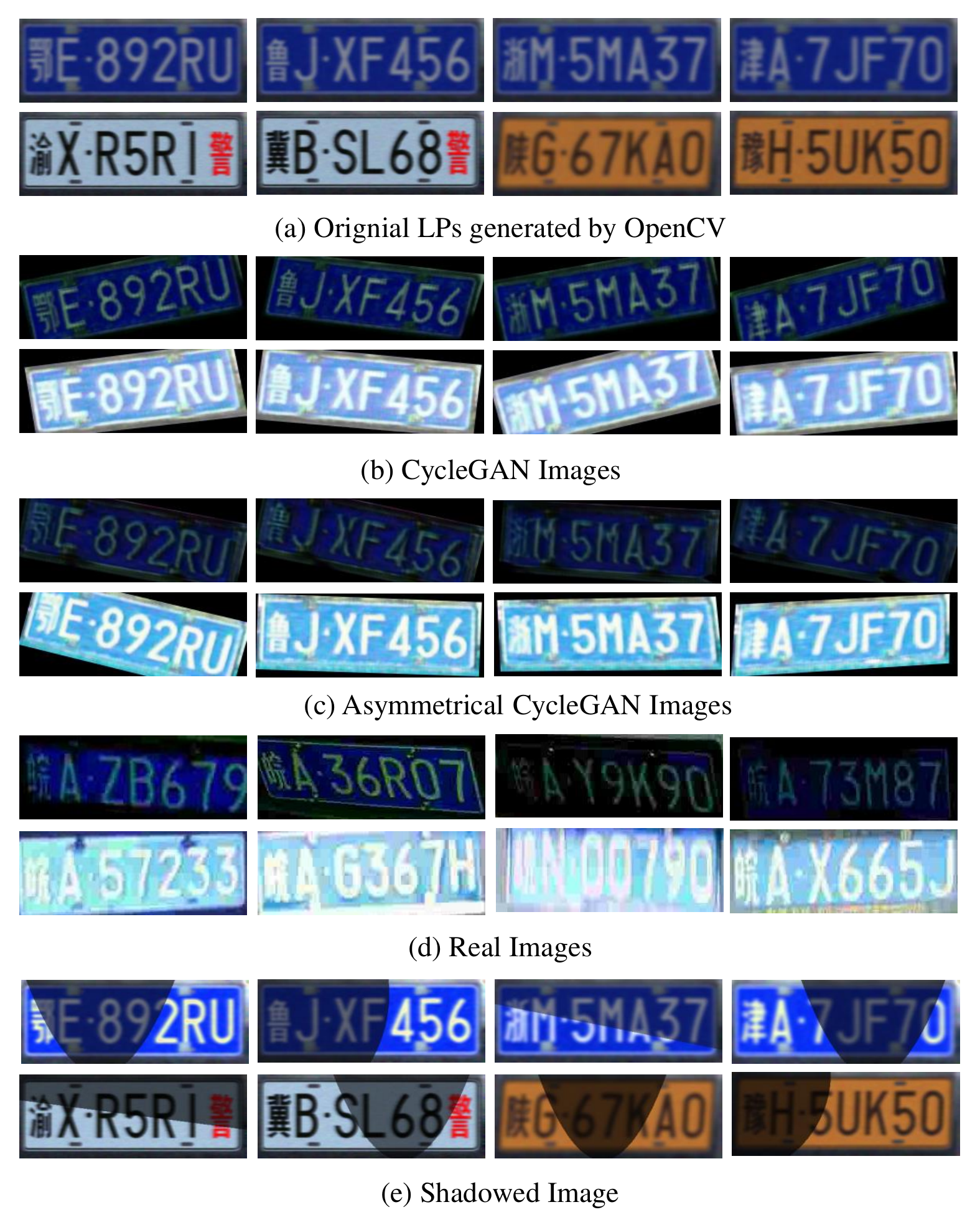}
\vspace{-3mm}
\caption{Various algorithms for generating license plate images. (a) Synthetic LPs generated by OpenCV; (b) The examples of LPs generated by CycleGAN model~\cite{Zhu2017Unpaired} ; (c) The examples of LPs generated by our asymmetric CycleGAN model; (d) Real LPs from CCPD-DB; (e) Shadowed Image.}
\label{compare}
\end{figure}

As in CycleGAN, the training of our proposed AsymCycleGAN model only requires two sets of unaligned images: synthetic and real images.
As shown in Figure~\ref{compare}, the synthetic images are generated using OpenCV, while the real images are sampled from the CCPD dataset~\cite{Xu2018Towards}.
To generate different types of real images, we further divide the CCPD images into two subsets with different illumination conditions: 
dark and bright. We use this dataset to train standard CycleGAN and our asymmetric CycleGAN model respectively, which consists of $800$ synthetic LPs generated by OpenCV and $800$ real-life license plate images in dark or glare environments.
The AsymCycleGAN model is trained with a learning rate of $0.0002$ and $30$ epochs.
The images generated by CycleGAN and asymmetric CycleGAN are shown in Figure~\ref{compare}. 
Moreover, we try to add shadows on the synthetic images so as to imitate real environment, the generated images are presented in Figure~\ref{compare} (e).

\section{The Proposed LP Dataset}
In this chapter, we introduce a new LP dataset named CLPD (China License Plate Dataset), 
for a more comprehensive evaluation of LP detection and recognition algorithms, including how it is collected  (Section~\ref{sec:datacl}) and the comparison with other datasets (Section~\ref{sec:datacomp}).
  
\subsection{Data Collection}
\label{sec:datacl}
\begin{figure}[t!]
\centering
\includegraphics[scale=0.31]{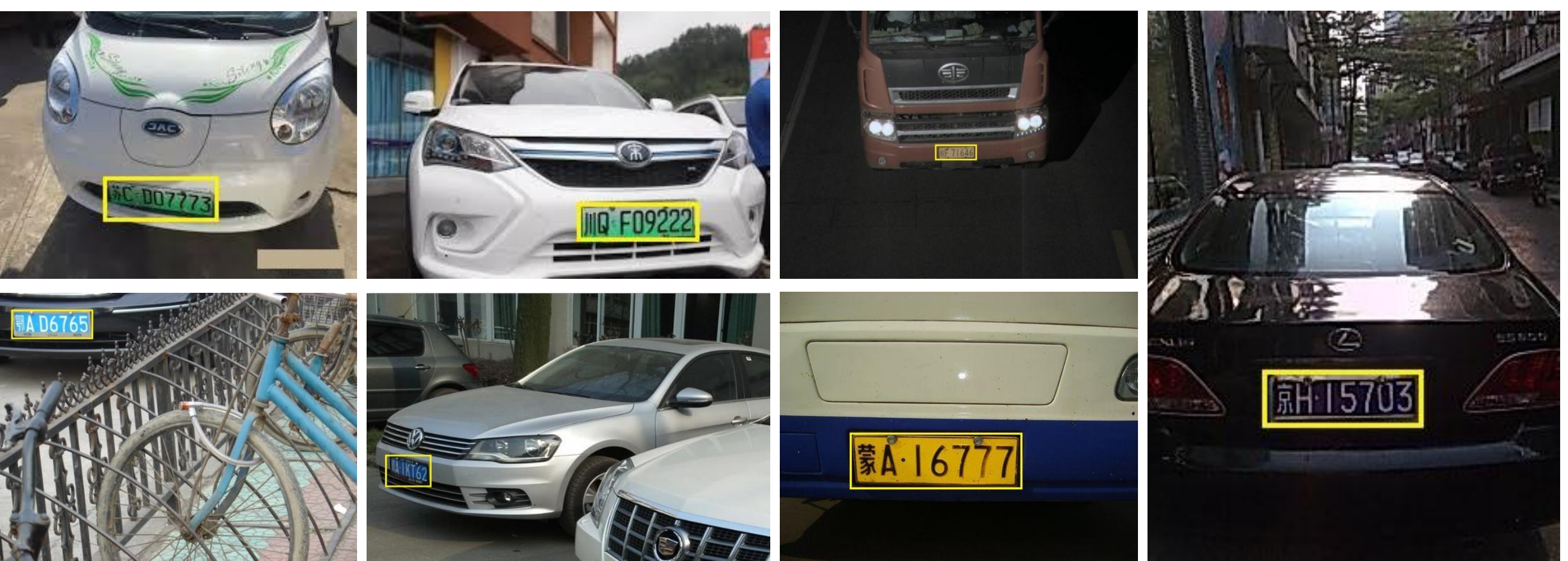}
\vspace{-5mm}
\caption{Sample images in our proposed CLPD dataset. Each license plate is manually annotated with a bounding box and its license number.} 
\label{CLPD}
\end{figure}

The LP images in the proposed CLPD dataset are collected from a variety of real-scene image sources, for example, searched from the Internet, taken by mobile phones or captured by car driving recorders.
All the faces shown in the images are blurred for privacy reasons.
When taking LP photos, we also diversify the photographing angles, shooting times, resolutions and background so as to cover different conditions.
The proposed dataset includes multiple vehicle types, such as trucks, cars, police cars and new energy vehicles.
Note that new energy vehicles in China have license plates with eight letters, while other vehicles have seven-letter license plates. We also allow occluded license plates which have less than seven visible letters. The variation in the length of license plate letters increases the recognition difficulty as well, and makes the rule based recognition methods infeasible. 
The bounding boxes and license plate letters are annotated manually. 
In summary, the CLPD dataset contains $1200$ LP images from all $31$ provinces in mainland China. Some examples are shown in Figure~\ref{CLPD}.
To our knowledge, our proposed LP dataset is the only one that covers all mainland China provinces with real shotted images.

\subsection{Dataset Comparison}
\label{sec:datacomp}


As presented in Table~\ref{datset_compare}, 
we compare our proposed dataset with other LP datasets in several aspects.
Although the size of our dataset is small, it contains the most number of region codes.
As we collect LP images from multiple sources, the image sizes are not fixed, in contrast to other datasets. 
Furthermore, AOLP, CCPD and our CLPD contain tilted images, while PKUData does not.
Finally, our dataset contains LPs from different types of vehicles, including police car, new energy car and truck, which further increases the diversity of LP styles.    

\begin{table}[t]
\centering
\caption{A comparison of available datasets for LP detection and recognition. LP size is the average size of all license plate areas in a dataset.}
\label{datset_compare}
\resizebox{0.49\textwidth}{!}{
\begin{tabular}{lcccc}
\hline
& AOLP~\cite{AOLP}& PKUData~\cite{PKUData} &  CCPD~\cite{Xu2018Towards}& CLPD (ours)       \\ \hline
Year                & $2012$         & $2016$              & $2018$       & $2019$  \\ \hline
\#Images            & $2049$         & $2253$               & $290k$       & $1200$       \\ \hline
\#Region Codes         & $0$            & $23$                  & $29$          & $31$       \\ \hline
LP size          & $72\times28$ &$156\times39$     & $253\times100$ & $149\times48$    \\ \hline
Tilted  & \checkmark & \ding{55} & \checkmark  & \checkmark \\ \hline
Var in vehicle type & \ding{55} & \checkmark & \ding{55}  & \checkmark \\ \hline
\end{tabular}}
\end{table}


\section{Experiments}
In this section, we conduct extensive experiments to compare our license plate recognition method with the state-of-the-art recognition methods. To demonstrate the effectiveness of the proposed model, plenty of experiments are performed on $4$ different license plate datasets.

\subsection{Datasets}

\textbf{CCPD}~\cite{Xu2018Towards} is currently the largest publicly available License Plate (LP) dataset that provides over $290k$ unique Chinese LP images with detailed annotations. This dataset is separated into different groups according to the difficulty of identification, the illuminations on LP area, the distance from the license plate when photographing, the degree of horizontal tilt and vertical tilt, and the weather (rainy, snowy or fog). Each category includes 10k to 20k images. CCPD-base consists of approximately $200k$ images, where $100k$ are used for training and the other half is for test. The other subdatasets (CCPD-DB, CCPD-FN, CCPD-Rotate, CCPD-Weather, CCPD-Challenge) are also used for test. 

\textbf{AOLP}\cite{AOLP} database consists of $2049$ images of Taiwan license plate. This dataset is categorized into three subsets according to complexity levels and photographing conditions: Access Control (AC), Traffic Law Enforcement (LE) and Road Patrol (RP). Since we do not have any other images with Taiwan license plate, we use any two of these subsets for training and the remaining one for test, similar to previous practices~\cite{LiReading2016,Hui2017Toward,wu2018many}.

\textbf{PKUData}\cite{PKUData} is released by Yuan \etal, which provides images for license plate detection. The license plate labels are not annotated and we labeled the $2253$ images in this dataset. {$1352$ images are randomly selected for training and the rest $901$ are used for test.}

\textbf{CLPD} is our proposed LP dataset, which contains $1200$ images across all provinces in mainland China, with different vehicle types included.  {The images in the newly proposed CLPD dataset are all real and cover a large variety of photographing conditions, vehicle types and region codes.
They are only used for test to verify the practicality of LP recognition models.
}

\subsection{Implementation Details}

In this work, we mainly focus on license plate recognition. In order to get the bounding boxes of license plates, a YOLOv2~\cite{DBLP:journals/corr/RedmonF16} detector is trained on the training set of CCPD. We set the IOU threshold to $0.6$, and achieve a detection performance of $precision=99.4\%$ and $recall=99.5\%$ on CCPD test sets. For fair comparison, we use the same evaluation criteria as that in~\cite{Xu2018Towards}. An LP recognition result is correct if and only if the IoU between the detection and the ground truth is greater than $0.6$ and all characters of the LP are correctly recognized (including the region code).

The recognition network is trained with cross-entropy loss and ADAM optimizer without any pre-training. In the training process, we adopt a batch size of $24$ and a learning rate of $1e^{-3}$ initially. The learning rate is multiplied by $0.9$ at every $12,000$ iterations until it reaches to $1e^{-5}$. The heights of input images in a batch are fixed, while the widths are calculated according to the aspect ratios of original images. All the experiments are conducted on an NVIDIA GTX1080Ti GPU with 11GB memory.

\subsection{Ablation Studies}

To analyze our proposed framework in detail, in this section, we evaluate it with different settings on CCPD dataset.

\subsubsection{Effect of CNN structures}
In order to analyze the impact of CNN capacities, we first experiment with different number of CNN channels and layers. 
As shown in Table~\ref{channel}, 
using more CNN channels indeed improves the license plate recognition accuracy, and the performance is saturated when the channel number reaches $512$.
Experimental results with different convolutional layers are demonstrated in Table~\ref{layers}.
The $30$-layer Xception performs better than models with less layers,
but the performance does not significantly improve when further increasing the depth.
Hereinafter, we use the $30$-layer Xception with $512$ channels.

\subsubsection{Effect of inaccurate bounding box}
Secondly, we test the recognition performance with detected and ground truth bounding boxes respectively, to demonstrate the robustness of our algorithm. Note that the detected bounding boxes may not encompass the license plates exactly as the groundtruth. This experiment is conducted to show the effect of bounding box variance on recognition performance.
As shown in Table~\ref{boundingbox}, the recognition accuracy only drops slightly by using detected bounding boxes (smaller than $0.2\%$ for all cases except the ``Challenge'' one), which validates the robustness of our algorithm to inaccurate bounding boxes.
One of the possible reasons is that the adopted 2D attention mechanism makes our algorithm not heavily depend on accurate bounding boxes: at each character decoding step, the adopted attention module will extract the most relevant local feature for each character in 2D space, instead of relying on heuristics rules for character separation.

\subsubsection{Effect of synthetic data}
The last ablation study is on the effectiveness of generated synthetic data. Here we also compare the performance by using different GAN models.
We train our model with different numbers of real and synthetic images ($20$k, $50$k and $100$k), and then test the performance on CCPD-DB dataset. 
As shown in Table~\ref{cycleGANcompare}, using the synthetic data generated by our proposed AsymCycleGAN offers better improvements than using that generated by the original CycleGAN, which demonstrates the superiority of our proposed AsymCycleGAN.

In addition, when comparing the improvements by using different number of real images, it can be found that the synthetic data plays a more important role when the real data size is smaller.

{
We can also see that the improvement is reduced when using smaller number of synthetic images.
Note that the cost of generating synthetic images is very cheap: they do not need human annotation 
and the generation speed is fast (about 1K/min).
So we can easily employ massive synthetic data for training, to improve the accuracy of LP recognition algorithms. 
}


\begin{table}[!t]
\centering\small
  \caption{Recognition accuracy ($\%$) with different CNN channels on CCPD subsets. $512$ channels will be adopted in furture experiments.}
  \label{channel}
  \resizebox{0.47\textwidth}{!}{
  \begin{tabular}{ccccccccc}
    \hline
    CNN Channels & Base & DB & FN & Rotate & Tilt & Weather & Challenge\\
    \hline \hline
    $128$  & $99.7$ & $98.6$ & $98.7$ & $96.1$ & $97.7$ & $98.1$ & $86.3$  \\
    \hline
    $256$  & $99.7$ & $98.9$ & $99.0$ & $97.1$ & $98.3$ & $98.2$ & $87.9$ \\
    \hline
    $512$  & $\mathbf{99.8}$ & $\mathbf{99.2}$ & $\mathbf{99.1}$ & $\mathbf{98.1}$ & $\mathbf{98.8}$ & $\mathbf{98.6}$ & $\mathbf{89.7}$ \\
    \hline
    $1024$  & $99.3$ & $98.0$ & $97.8$ & $94.2$ & $96.4$ & $97.4$ & $83.4$ \\
    \hline
  \end{tabular}}
\end{table}

\begin{table}[!t]
\caption{Comparing the recognition accuracy ($\%$) with different layers of Xception on CCPD subsets. We choose the optimal 30-layers Xception for feature extraction.}
\label{layers}
\resizebox{0.48\textwidth}{!}{
  \begin{tabular}{ccccccccc}
    \hline
    Layers & Base & DB & FN & Rotate & Tilt & Weather & Challenge\\
    \hline \hline
    $15$  & $98.8$ & $98.6$ & $98.4$ & $96.4$ & $97.9$ & $97.8$ & $87.6$ \\
    \hline
    $20$  & $99.2$ & $98.8$ & $98.7$ & $97.1$ & $98.4$ & $98.3$ & $88.6$ \\
    \hline
    $25$  & $99.6$ & $99.0$ & $98.9$ & $97.6$ & $98.5$ & $98.6$ & $89.0$ \\
    \hline
    $30$  & $\mathbf{99.8}$ & $\mathbf{99.2}$ & $99.1$          & $\mathbf{98.1}$ & $\mathbf{98.8}$ & $\mathbf{98.6}$ & $\mathbf{89.7}$ \\
    \hline
    $35$  & $\mathbf{99.8}$ & $99.1$          & $99.1$          & $\mathbf{98.1}$ & $98.7$          & $\mathbf{98.6}$ & $89.2$ \\
    \hline
    $40$  & $\mathbf{99.8}$ & $\mathbf{99.2}$ & $\mathbf{99.2}$ & $98.0$          & $98.7$          & $98.5$          & $89.4$ \\
    \hline
  \end{tabular}}
\end{table}

\begin{table}[!t]
\centering\small
  \caption{Recognition accuracy ($\%$) by using different bounding boxes on sub-datasets of CCPD. The experimental results show small gap when using inaccurate bounding boxes, which demonstrates the robustness of our algorithm.}
  \label{boundingbox}
  \resizebox{0.48\textwidth}{!}{
  \begin{tabular}{ccccccccc}
    \hline
    Bounding Box & Base & DB & FN & Rotate & Tilt & Weather & Challenge\\
    \hline \hline
    by Detection & $99.8$ & $99.2$ & $99.1$ & $98.1$ & $98.8$ & $98.6$ & $89.7$ \\
    \hline
    Ground truth & $\mathbf{99.8}$ & $\mathbf{99.4}$ & $\mathbf{99.3}$ & $\mathbf{98.2}$ & $\mathbf{98.9}$ & $\mathbf{98.7}$ & $\mathbf{90.1}$ \\
    \hline
  \end{tabular}}
\end{table}

\begin{table}[!t]
	\small
	\centering
	\caption{The recognition accuracy ($\%$) on CCPD-DB, with different number of real images and different GAN models adopted. Using synthetic data generated by our proposed AsymCycleGAN offers better performance. The superiority is even obvious if there are a small number of real images.}
	\label{cycleGANcompare}
	\resizebox{0.48\textwidth}{!}{
			\begin{tabular}{l|l|l}
			\hline
			Training Data                    &  \multicolumn{1}{c|}{CCPD-DB}      & \multicolumn{1}{c}{Improvement}\\ \hline \hline
			Real (20k)                       &  \multicolumn{1}{c|}{$96.1$}  & \multicolumn{1}{c}{}\\ \hline 
			{Real (20k) + CycleGAN (20k)} &  \multicolumn{1}{c|}{{$96.3$}}  & \multicolumn{1}{c}{{$0.2$}}\\ \hline
			{Real (20k) + AsymCycleGAN (20k)}  &  \multicolumn{1}{c|}{{$96.3$}}  & \multicolumn{1}{c}{{$0.2$}}\\ \hline  
			Real (20k) + CycleGAN (200k)     &  \multicolumn{1}{c|}{$96.6$}  & \multicolumn{1}{c}{$0.5$}\\ \hline
			Real (20k) + AsymCycleGAN (200k) &  \multicolumn{1}{c|}{$\mathbf{96.8}$}  & \multicolumn{1}{c}{$\mathbf{0.7}$}\\ \hline \hline

			Real (50k)                        &  \multicolumn{1}{c|}{$97.9$}  & \multicolumn{1}{c}{} \\ \hline
			{Real (50k) + CycleGAN (50k)}&  \multicolumn{1}{c|}{{$98.0$}}  & \multicolumn{1}{c}{{$0.1$}}\\ \hline
			{Real (50k) + AsymCycleGAN (50k)}&  \multicolumn{1}{c|}{{${98.1}$}}  & \multicolumn{1}{c}{{${0.2}$}}\\ \hline  
			Real (50k) + CycleGAN (200k)      &  \multicolumn{1}{c|}{$98.3$}  & \multicolumn{1}{c}{$0.4$}\\ \hline
			Real (50k) + AsymCycleGAN (200k)  &  \multicolumn{1}{c|}{$\mathbf{98.4}$}  & \multicolumn{1}{c}{$\mathbf{0.5}$}\\ \hline  \hline

			Real (100k)                       &  \multicolumn{1}{c|}{$98.8$}  & \multicolumn{1}{c}{} \\ \hline
			{Real (100k) + CycleGAN (100k)}     &  \multicolumn{1}{c|}{{$98.8$}}  & \multicolumn{1}{c}{{$0.0$}} \\ \hline
			{Real (100k) + AsymCycleGAN (100k)} &  \multicolumn{1}{c|}{{$98.8$}}  & \multicolumn{1}{c}{{$0.0$}}\\ \hline
			Real (100k) + CycleGAN (200k)     &  \multicolumn{1}{c|}{$98.9$}  & \multicolumn{1}{c}{$0.1$} \\ \hline
			Real (100k) + AsymCycleGAN (200k) &  \multicolumn{1}{c|}{$\mathbf{99.0}$}   & \multicolumn{1}{c}{$\mathbf{0.2}$}\\ \hline
	\end{tabular}}
\end{table}

\subsection{Experiments on Existing Benchmarks}

\subsubsection{Results on CCPD}

\begin{table*}[!t]
\centering\small
  \caption{LP recognition accuracy ($\%$) on each CCPD test set (Number of images in parentheses). We achieve the highest recognition accuracy compared with other algorithms, especially in the datasets with rotattion and challenging license plates.}
  \label{accuracy}
  \resizebox{0.82\textwidth}{!}{
  \begin{tabular}{lcccccccccc}
    \hline
    Model & Overall & Base & DB & FN & Rotate & Tilt & Weather & Challenge & {Test time}\\
    \#Images& &($100k$)&($20k$)&($20k$)&($10k$)&($10k$)&($10k$)&($10k$)&{{ms}}\\
    \hline
    Ren \textsl{et al.} (2015) \cite{Ren2015Faster} & $92.8$ & $97.2$ & $94.4$ & $90.9$ & $82.9$ & $87.3$ & $85.5$ & $76.3$ & {$57.6$}\\
    Liu \textsl{et al.} (2016)  \cite{Liu2016SSD} & $95.2$ & $98.3$ & $96.6$ & $95.9$ & $88.4$ & $91.5$ & $87.3$ & $83.8$ & {$25.6$}\\
    Joseph \textsl{et al.} (2016) \cite{DBLP:journals/corr/RedmonF16} & $93.7$ & $98.1$ & $96.0$ & $88.2$ & $84.5$ & $88.5$ & $87.0$ & $80.5$ & {$23.8$}\\
    Li \textsl{et al.} (2017) \cite{Hui2017Toward} & $94.4$ & $97.8$ & $94.8$ & $94.5$ & $87.9$ & $92.1$ & $86.8$ & $81.2$ & {$310$}\\
    
    Zherzdev \textsl{et al.} (2018) \cite{zherzdev2018lprnet} & $93.0$ & $97.8$ & $92.2$ & $91.9$ & $79.4$ & $85.8$ & $92.0$ & $69.8$ & {$17.8$}\\
    
    Xu \textsl{et al.} (2018) \cite{Xu2018Towards} & $95.5$ & $98.5$ & $96.9$ & $94.3$ & $90.8$ & $92.5$ & $87.9$ & $85.1$ & {$\mathbf{11.7}$}\\

    {Zhang \textsl{et al.} (2019) \cite{zhang2016joint, zherzdev2018lprnet} }& {$93.0$ }&{ $99.1$ }& {$96.3$ }&{$97.3$} &{ $95.1$ }& {$96.4$ }&{ $97.1$ }&{ $83.2$} & {$153$}\\
    
    {Luo \textsl{et al.} (2019) \cite{luo2019moran} }&{ $98.3$ }&{ $99.5$ }&{ $98.1$ }& {$98.6$} &{ $98.1$} &{ $98.6$ }& {$97.6$} & {$86.5$} & {$18.2$}\\
    
    {Wang \textsl{et al.} (2020) \cite{wang2019decoupled}} & {$96.6$} & {$98.9$} & {$96.1$} & {$96.4$} & {$91.9$} & {$93.7$} & {$95.4$} & {$83.1$} & {$19.3$}\\
    
    \hline
    Ours (Real Data Only)     & $98.5$ & $99.6$ &  $98.8$ & $98.8$ & $96.4$ & $97.6$ & $98.5$ & $88.9$ & {$24.9$}\\
    Ours (Real + Synthetic data) & $\mathbf{98.9}$ & $\mathbf{99.8}$ & $\mathbf{99.2}$ & $\mathbf{99.1}$ & $\mathbf{98.1}$ & $\mathbf{98.8}$ & $\mathbf{98.6}$ & $\mathbf{89.7}$\\
    \hline
  \end{tabular}}
\end{table*}


{
It can be seen from Table~\ref{accuracy} that 
our algorithm outperforms other algorithms in terms of the overall AP and most of subsets, using the same real training data.
The only exception is that the method of Luo \textsl{et al.} \cite{luo2019moran} is better than ours on the rotate and tilt subsets. 
The reason may be that Luo \textsl{et al.} \cite{luo2019moran} adopts an STN-based~\cite{jaderberg2015spatial} technique which is specifically designed for rotated images. 
Note that our algorithm can also benefit from using this technique and the accuracy on the rotate and tile subset is expect to be further improved. 
}

Our algorithm shows significant superiority on subsets with irregular LP images, such as ``Rotate'', ``Weather'' and ``Challenge'', which again proves the robustness of our model to the deformation of license plates. Moreover, by adding synthetic images generated by our AsymCycleGAN, the recognition accuracies consistently raise furthermore on all subsets (a $0.4\%$ gain of Overall). 
The increment is even obvious when LPs are rotated or tilted (raising $1.7\%$ on CCPD-Rotate and $1.2\%$ on CCPD-Tilt).
The main reason is that random perspective transformation and rotation are applied to the synthesized data, which is a great complementary to real data.

\begin{figure*}[!t]
\centering
\includegraphics[scale=0.50]{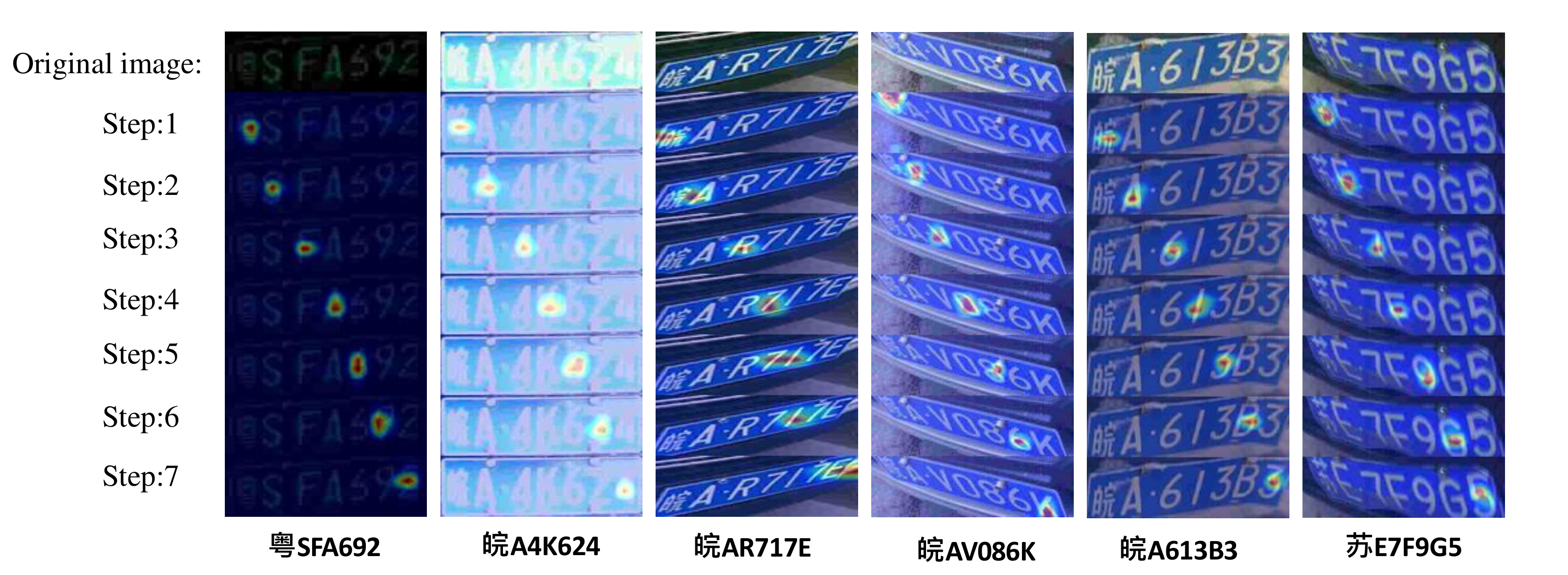}
\caption{{Visualization of 2D attention weights at each decoding timestep. Results indicate that the 2D-attention model can handle challenging cases.}} 
\label{heatmap}
\end{figure*}

We select some extremely distorted images and visualize the 2D attention heat maps when decoding each character in Figure \ref{heatmap}. The results show that even for very tilted images, the 2D attention model can locate to the character being decoded and extract corresponding features for recognition. It should be noted that the attention module does not require additional character-level annotations. It is trained in a weakly supervised manner by the cross-entropy loss on the whole plate recognition.

\subsubsection{Results on AOLP}

\begin{table}[!t]
\centering
  \caption{The recognition accuracy ($\%$) on sub-datasets of AOLP. Our approach performs better than other methods on all three subsets.}
  \label{res_AOLP}
  \begin{tabular}{lcccccccc}
    \hline
    Model\quad & \quad AC & \quad LE & \quad RP \quad\\
    \#Images \quad & \quad (681)& \quad (757)& \quad (611) \quad\\
    \hline
    Li \textsl{et al.} (2016) \cite{LiReading2016} \quad & \quad $94.9$  & \quad $94.2$  \quad & \quad $88.4$ \quad \\ \hline
    Li \textsl{et al.} (2017) \cite{Hui2017Toward} \quad & \quad $95.3$  & \quad $96.6$  \quad & \quad $83.7$ \quad \\ \hline
    Wu \textsl{et al.} (2018) \cite{wu2018many}    \quad & \quad $96.6$  & \quad $97.8$  \quad & \quad $91.0$ \qquad \\ \hline
    Ours  \quad &  \quad {$\mathbf{97.3}$}  &\quad  {$\mathbf{98.3}$} \quad & \quad {$\mathbf{91.9}$} \quad\\
    \hline
  \end{tabular}
\end{table}

In this section, we compare our model with other state-of-the-art methods on AOLP dataset. For fair comparison, we did not use any synthetic data during model training. Perspective transformation is employed for data augmentation.
The results in Table~\ref{res_AOLP} show that our approach performs better than other methods on all three subsets, which validates the superiority of our approach. In particular, our method leads to the accuracy increments of $0.7 \%$ on AC, $0.5 \%$ on LE and $0.9 \%$ on RP, compared to the second best results. Note that the RP subset is mainly composed of oriented or distorted license plates, on which our method obtains the largest performance gain. This result further demonstrates the effectiveness of our model in recognizing irregular license plates.




\subsubsection{Results on PKUData}
For PKUData, we randomly sample three-fifths for training and use the remaining two-fifths for test. For fair comparison, we re-train the model proposed in~\cite{Xu2018Towards} by the same training data.
An open API called Sighthounds~\cite{Masood2017} is tested as well, but we have no idea about the training data it used.
We evaluate the LP recognition accuracy on two settings, \ie, with and without region code (a Chinese character) considered. The model in Sighthounds~\cite{Masood2017} does not support region code recognition, so we only report its accuracy without region code.
The recognition results are shown in Table~\ref{res_PKUData}. Our model outperforms that in~\cite{Xu2018Towards} by about $7 \%$ when only real data is adopted, and surpasses Sighthounds~\cite{Masood2017} if synthetic training data is added. In comparison with the improvement on CCPD dataset, the accuracy gain is even more obvious when using synthetic data (about $4 \%$), because of the limited real training images in PKUData, which demonstrates the usefulness of our synthesis engine when there is scarce training data.


\begin{table}[!t]
\centering
\caption{The recognition accuracy (ACC,$\%$) and recognition accuracy without region code (ACC w/o RC,$\%$) on PKUData and CLPD. For ACC w/o RC, the recognition is considered to be correct if all the characters except the first one region code are correctly recognized.}
\label{res_PKUData}
\resizebox{0.48\textwidth}{!}{
\begin{tabular}{l|c|c|c|c}
\hline
Dataset     & \multicolumn{2}{c|}{PKUData } & \multicolumn{2}{c}{ CLPD } \\ \hline
Criterion &  ACC   & ACC w/o RC &  ACC  & ACC w/oRC  \\ \hline
Masood \textsl{et al}. (2017)~\cite{Masood2017} & \multicolumn{1}{c|}{-} &\multicolumn{1}{c|}{$89.3$} &\multicolumn{1}{c|}{-} &\multicolumn{1}{c}{$85.2$}\\ \hline
Xu \textsl{et al}. (2018) \cite{Xu2018Towards} &\multicolumn{1}{c|}{$77.6$} &\multicolumn{1}{c|}{$78.4$} &\multicolumn{1}{c|}{$66.5$}& \multicolumn{1}{c}{$78.9$}\\\hline
Ours (Real Data Only)&\multicolumn{1}{c|}{$84.8$}&\multicolumn{1}{c|}{$86.5$}&\multicolumn{1}{c|}{$70.8$}&\multicolumn{1}{c}{$86.1$}\\ \hline
Ours (Real + Synthetic Data)&\multicolumn{1}{c|}{$\mathbf{88.2}$}&\multicolumn{1}{c|}{$\mathbf{90.5}$}&\multicolumn{1}{c|}{$\mathbf{76.8}$}&\multicolumn{1}{c}{$\mathbf{87.6}$}\\ \hline
\end{tabular}}
\end{table}

\subsection{Experiments on Our CLPD Dataset}
\begin{figure}[!t]
\centering
\includegraphics[scale=0.34]{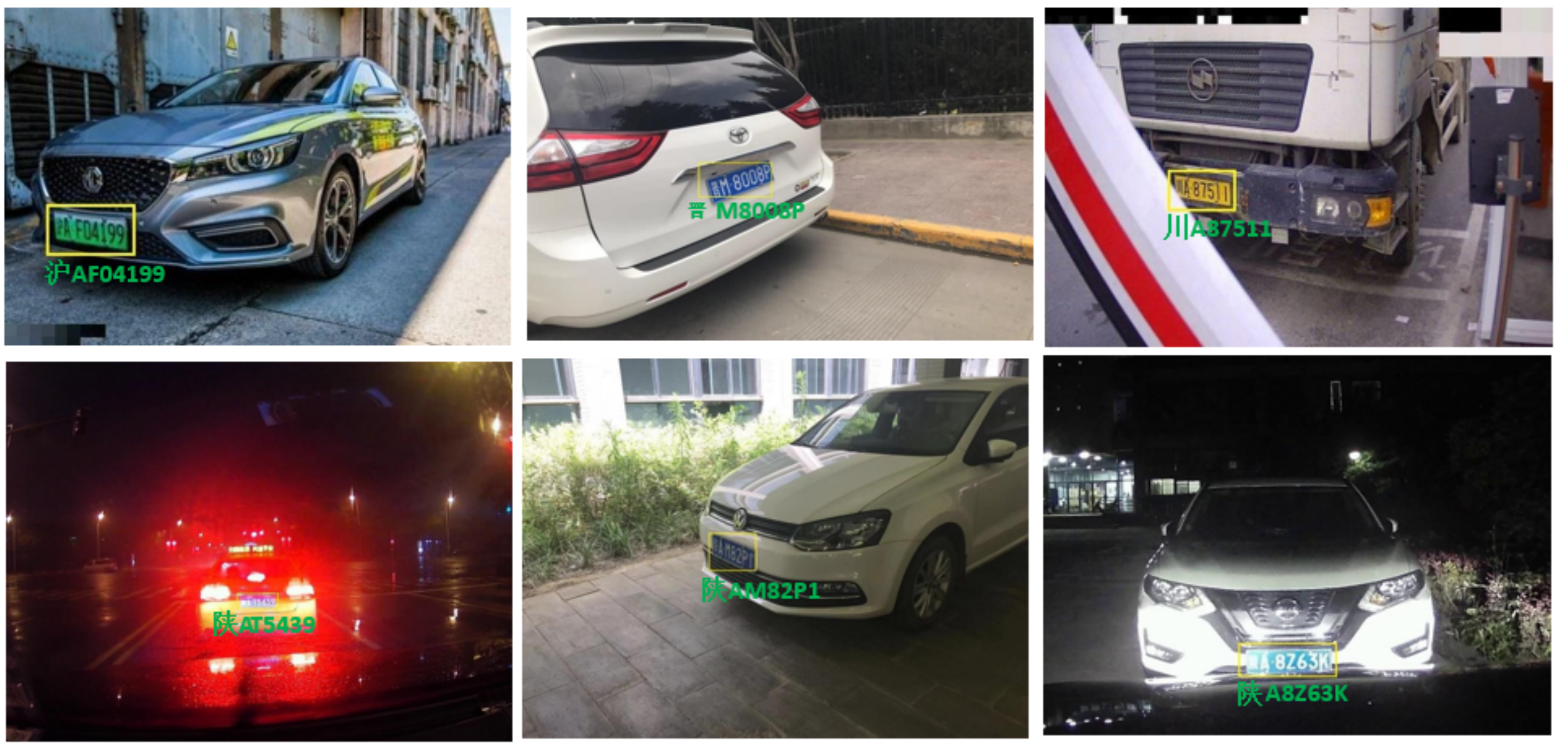}
\vspace{-5mm}
\caption{Detection and recognition results on CLPD using YOLOv2 and our recognition model. With the addition of synthetic data, the model is able to recognize license plates from different provinces under various scenarios.} 
\label{CLPD_result}
\end{figure}

As aforementioned, the diversity of our proposed CLPD dataset is much larger than 
existing LP datasets, which provides a platform to evaluate current algorithms comprehensively.
We train the proposed model on CCPD-Base dataset, and test it on CLPD. Experimental results in Table~\ref{res_PKUData} show the advantage of our model. It leads to the highest accuracy no matter region code is considered or not. By adding synthetic data, the accuracy increases $6 \%$ further if region code is considered, which benefits from a more balanced region code distribution in our synthetic data that can be easily obtained by the proposed engine. Some experimental results are visualized in Figure~\ref{CLPD_result}.

\begin{figure}[!t]
\centering
\includegraphics[scale=0.52]{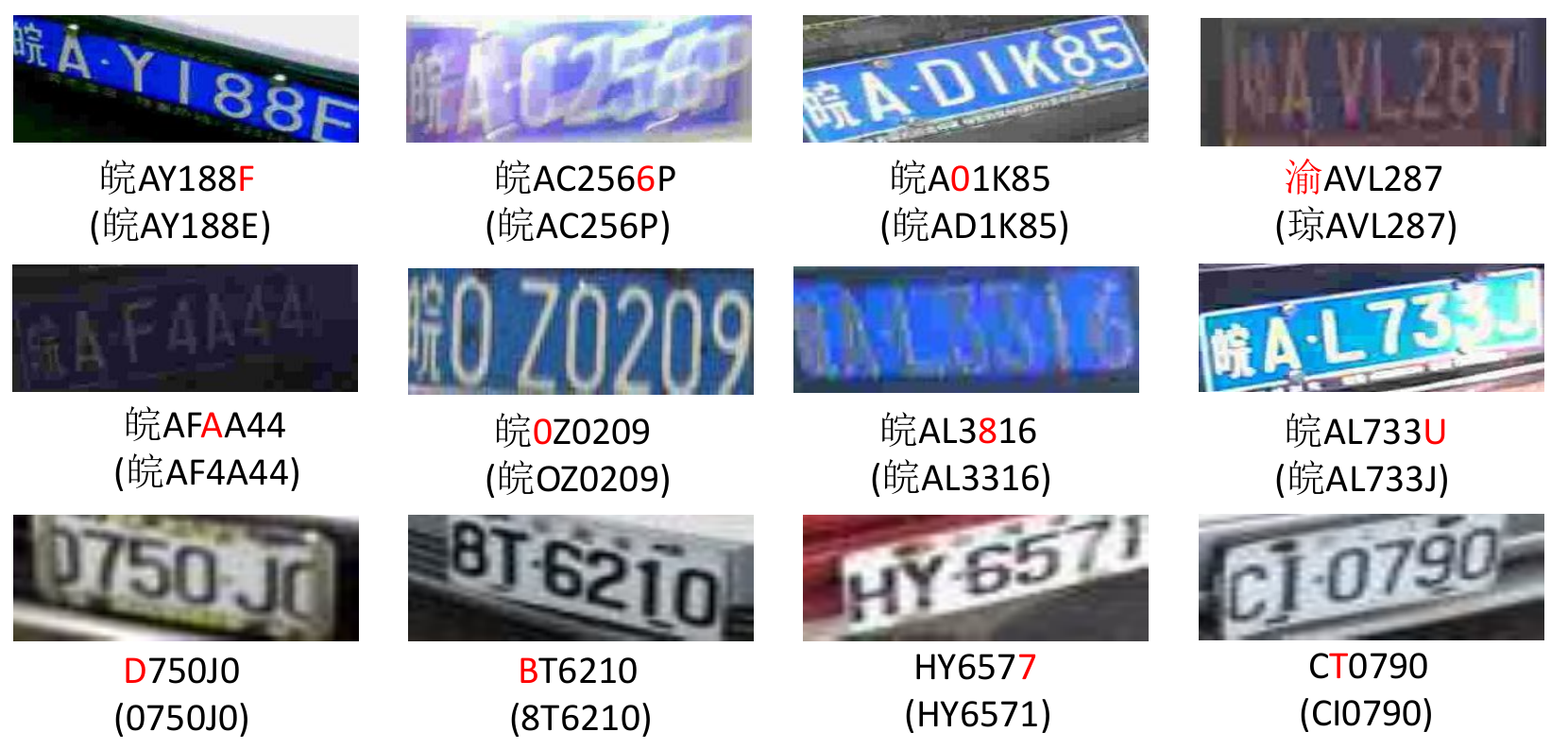}
\vspace{-5mm}
\caption{Examples of LPs that are incorrectly recognized by the proposed method. The ground truth is shown in the parenthese.} 
\label{error}
\end{figure}

We also present some failure cases in Figure~\ref{error}. As there is no specific language rule used in license plate, some similar characters are rather difficult to be distinguished, such as ``4'' and ``A'', ``8'' and ``B'',``0'' ``D'', and ``O''. Images with extreme blur or occlusion are also unable to be recognized.



\section{Conclusion}                  

In this paper, we present a robust model for license plate recognition in unconstrained environment. The proposed model is built upon an Xception CNN module for feature extraction, and a 2D-attention based RNN module for sequence decoding. To handle the shortage or unbalance of real training data, CycleGAN is tailored to generate synthetic LP images with different deformation styles and a more balanced region codes, which provides a simple yet effective way to complement available real data. Extensive experimental results indicate the superiority of our methods, especially when addressing distorted license plates or with limited training data. An LP dataset that contains images captured in different ways from various regions is collected so as to evaluate LP recognition methods more comprehensively.  

We use an LSTM-based sequence decoder for license plate recognition, which cannot be trained in parallel over time steps. For future works, a transformer-like decoder may be explored to accelerate training speed.


{\small

	\bibliographystyle{ieee}
}
\end{document}